\documentclass[12pt]{article}

\usepackage[colorlinks,urlcolor=blue,linkcolor=blue,citecolor=blue]{hyperref}
\usepackage{color,array}
\usepackage{graphicx}
\usepackage{multirow}
\usepackage[vlined,ruled,linesnumbered]{algorithm2e}
\usepackage{amsfonts}
\usepackage{amssymb}
\usepackage{amsmath}
\usepackage{bbm}
\usepackage{fancyhdr}
\usepackage{cancel}

\newcommand{\omitit}[1]{}

\SetKwRepeat{Do}{do}{while} % for using while loop inside algorithm

\fancypagestyle{fancyfirst}
{
\fancyhead{}
\fancyfoot[C]{
\footnotesize DISTRIBUTION STATEMENT A. Approved for public release. Distribution unlimited.\\
This work has been submitted to the IEEE for possible publication. Copyright may be transferred without notice, after which this version may no longer be accessible.
}
}

\begin{document}

\newcounter{defcounter}
\setcounter{defcounter}{1}

\newcounter{thmcounter}
\setcounter{thmcounter}{1}

\title{Divide and Repair: Using Options to Improve Performance of Imitation Learning Against Adversarial Demonstrations} 

\author{Prithviraj Dasgupta\\[0.02in]
Distributed Intelligent Systems Section\\
Information Technology Division\\
Naval Research Laboratory, Washington, D. C., USA \\
E-mail: raj.dasgupta@nrl.navy.mil
}

\maketitle

\thispagestyle{fancyfirst}
\begin{center}
{\bf Abstract}\\
\end{center}
We consider the problem of learning to perform a task from demonstrations given by teachers or experts, when some of the experts' demonstrations might be adversarial and demonstrate an incorrect way to perform the task. We propose a novel technique that can identify parts of demonstrated trajectories that have not been significantly modified by the adversary and utilize them for learning, using temporally extended policies or options. We first define a trajectory divergence measure based on the spatial and temporal features of demonstrated trajectories to detect and discard parts of the trajectories that have been significantly modified by an adversarial expert, and could degrade the learner's performance, if used for learning, We then use an options-based algorithm that partitions trajectories and learns only from the parts of trajectories that have been determined as admissible. We provide theoretical results of our technique to show that repairing partial trajectories improves the sample efficiency of the demonstrations without degrading the learner's performance. We then evaluate the proposed algorithm for learning to play an Atari-like computer-based game called {\tt LunarLander} in the presence of different types and degrees of adversarial attacks of demonstrated trajectories. Our experimental results show that our technique can identify adversarially modified parts of the demonstrated trajectories and successfully prevent the learning performance from degrading due to adversarial demonstrations.
%without introducing a significant computational or time overhead for the options-related calculations.

\section{Introduction}
%Behavior cloning is a fundamental part of reinforcement learning (RL) algorithms where an expert demonstrates a sequence of actions corresponding to expert behavior to the learning agent. These actions and corresponding states are used to initialize rewards at the state-action pairs reached by the agent and are crucial to the policy learned by the learning agent using RL. Clearly, if the expert demonstrations are corrupted, the learning agent perceives incorrect demonstrations, for instance, an incorrect pairings between actions and rewards, it ends up learning an incorrect policy.
Learning from demonstrations is a widely-used form of machine learning where a teacher or expert provides demonstrations of how to perform the learning task to speed up the learning process~\cite{hussein2017imitation, ravichandar2020recent} in the context of reinforcement learning~\cite{SuttonB18}. It has been used in many successful applications of machine learning algorithms including autonomous driving~\cite{pan2020imitation}, robotic manipulation~\cite{fang2019survey}, and human-robot interaction~\cite{mukherjee2022survey}. Conventionally, the experts demonstrating the task are assumed to be benign and show the correct way of performing the task. However, as machine learning-based autonomous systems become more pervasive, they are exposed to demonstrations from a variety of sources. Some of these demonstrations might be from adversarial experts that give incorrect demonstrations with the intention of making the autonomous system behave in incorrect and unintended ways. To address this problem, researchers have developed techniques for learning reliably in the presence of adversarial expert demonstrations~\cite{gleave2019adversarial, chen2022anomaly}. The main idea in most of these techniques is to use an eligibility metric, such as a confidence measure, on trajectories or temporal sequences of state-action pairs representing expert demonstrations, followed by accepting or rejecting the trajectories based on that metric. These techniques work with full or end-to-end (initial state to final state) trajectories, that is, the eligibility metric is calculated for the full trajectory, and, if found ineligible, the full trajectory is discarded. In this paper, we posit that even though the full trajectory might cause the learning task to fail, there could be parts of the trajectory that were benign, possibly show a new way of performing a part of the task, and could benefit the learning process. This insight is based on the observation that many adversarial attacks on machine learning algorithms are composed by modifying the input (e.g., training data examples for supervised learning~\cite{huang2017adversarial} or demonstrated trajectories for reinforcement learning~\cite{mandlekar2017adversarially}) only at certain, strategic features or locations, instead of all across the input. To address the problem of adversarial learning from demonstrated trajectories while retaining usable parts of the trajectories, we propose a novel technique using temporally extended policies or options~\cite{sutton1999between}. Our technique consists of two steps: first, we develop a divergence measure that can indicate the degree of deviation in expert demonstrations with respect to a small set of demonstrations that are guaranteed to be benign. We then use options to partition demonstrated trajectories and use the divergence measure to selectively accept or discard parts of demonstrated trajectories. We have provided theoretical analyses to show that our proposed technique of accepting only non-adversarial portions of trajectories for learning can prevent degrading of the learner's performance. We have also validated the technique using different types and degrees of attacks made by an adversary while learning to play an Atari-like game called {\tt LunarLander} using a form of learning from demonstrations called imitation learning. Our results show that our proposed technique can be used to identify and learn only from acceptable parts of demonstrated trajectories to improve the rewards from imitation learning in the presence of adversarial demonstrations. To the best of our knowledge, our work is one of the first attempts at integrating divergence measure with options to address the problem of adversarial learning from demonstrations.

The rest of this paper is structured as follows: in the next section, we provide an overview of relevant literature in adversarial reinforcement learning focusing on imitation learning. We then introduce the mathematical framework for the problem, measures for characterizing demonstrations given in the form of trajectories, and our option-based algorithm for partitioning and using acceptable parts of demonstrated trajectories for imitation learning. Sections~\ref{sec:theoritical_analysis} and ~\ref{sec:expts} provide the theoretical and experimental evaluation results of our proposed techniques, respectively, and, finally, we conclude. A preliminary version of this research is in~\cite{dasgupta21options}. In this paper, we have thoroughly rewritten the paper, formalized the mathematical framework, proposed new algorithms, and added new theoretical and experimental results.

\section{Related Work}
Adversarial learning has gained prominence over the past decade as an essential means to guarantee desired behavior of machine learning-based systems deployed in the real world. Here we discuss relevant literature on adversarial reinforcement learning (RL); a comprehensive survey of adversarial supervised learning is in~\cite{Chakraborty2018AdversarialAA}. 

Early researchers considered adversarial RL in the context of an RL agent learning suitable actions to play a competitive game like keep-away soccer against a player called an adversary~\cite{uther1997adversarial}, where the adversary's intent was to defeat the RL agent, albeit via fair play instead of using malicious tactics such as incorrect actions to misguide the RL agent. Subsequently, researchers proposed techniques where the expert demonstrator modifies the trajectory it demonstrates either indirectly or directly. In the former direction, researchers have considered including a risk term representing the demonstrator's possible deviations from optimal trajectories inside the Q-value function used by an RL agent to determine its policy ~\cite{pan2019risk}. In the latter direction, Mandlekar {\em et al.}~\cite{mandlekar2017adversarially} proposed a technique where the demonstrator directly modifies a valid trajectory using a perturbation technique like fast gradient sign method (FGSM)~\cite{goodfellow2014explaining} to create adversarial trajectories that are then demonstrated to the RL agent. The RL agent trains with both clean and adversarial demonstrations so that the learned policy can perform effectively even in the presence of adversarial demonstrations. Our work in this paper is complementary to this research and investigates options as a means to improve the rewards received by an RL agent in the presence of adversarial trajectory demonstrations.

Recently, authors~\cite{PintoDSG17}, ~\cite{gleave2019adversarial} have also investigated adversarial RL as a competitive zero-sum game where an adversarial demonstrator and an RL agent interact with each other but the learning objectives of the demonstrator and the RL agent are known to be directly contradictory to each other. Experimental results with simulated demonstrations of body movements on robotic figures showed that the demonstrator could successfully use its calculated policies to determine actions that misguided the RL agent to learn incorrect actions and lose stability instead of learning its intended task like walking or kicking a ball. In contrast to these scenarios where the demonstrator explicitly reveals its motive to make the RL agent fail by selecting incorrect states and actions, our research considers a more practical scenario where the demonstrator tries to stealthily modify some demonstrations that could make the RL agent fail, without revealing the demonstrators adversarial motives to the RL agent.

%In~\cite{PanSGC19}, authors add the variance in the Q-values as a risk term to the Q-value function of a Q-learning-based, learning agent to model the uncertainty in the Q-values induced by an adversary. ~\cite{PintoDSG17} consider a zero-sum game where the objective of the adversary is to find high reward trajectories for itself (and, consequenly, low reward trajectories for the learning agent, due to the zero-sum nature of the game). The learning agent and adversary alternatively update their policy parameters until convergence to the min-max outcome (or, Nash equilibrium) is reached. However, in many real-life settings, a zero-sum-like correlation between the adversary and learner agent's reward functions might not be valid, and, more general techniques instead of solving a min-max problem are required

\omitit{Recently, similar to some of the findings in our paper, authors~\cite{pan2022characterizing} have demonstrated that adversarial attacks on reinforcement learning results in reduced rewards to the learning agent in simulated and physical robotic mobility tasks.}

Another direction on adversarial RL integrated the techniques of inverse reinforcement learning~\cite{abs-1806-06877} and generative adversarial networks~\cite{goodfellow2014generative} in the generative adversarial imitation learning (GAIL) framework~\cite{ho2016generative} where trajectories are generated both by an expert using the expert policy, and, by a generator using the policy being learned. A discriminator evaluates the source of these trajectories and the learned policy is deemed as converged when the discriminator is unable to distinguish whether the trajectory was generated by the expert versus the generator. GAIL has also been extended to state-only observations with minimum demonstrations using sparse action guided regularization~\cite{sun2019adversarial}, and, to generative adversarial imitation from observation (GAIfO) that uses a cost function that depends on state observations only~\cite{TorabiWS19a}. The main difference between our work and GAIL is that, whereas, in GAIL the adversary or generator's objective is to update the policy being learned for faster convergence to an optimal policy. our work considers that the adversary's objective is to demonstrate incorrect trajectories to misguide the learner that is learning the policy. 

Our research is closely related to techniques for imitation learning with imperfect expert demonstrations. In these techniques, a rank or confidence score for each trajectory is provided either as input or via learning. This score is then used to update the trajectory's rewards and selectively include the trajectory in the training during learning. In~\cite{brown2019extrapolating, brown2020better} trajectories are associated with a score or rank that is provided as input or self-generated and used to revise the rewards of sub-optimal trajectories using inverse reinforcement learning. In~\cite{wu2019imitation}, authors proposed techniques called 2IWIL and ICGAIL that use semi-supervised inverse reinforcement learning techniques to calculate a confidence score for unlabeled trajectories while using a small set of confidence score-labeled trajectories. An inverse dynamics function is learned in~\cite{cao2021learning} to calculate a transformed trajectory from each expert trajectory followed by using a distance measure between the expert and transformed trajectory to determine a feasibility score for the expert trajectory. In~\cite{chen2022anomaly}, anomaly detection between trajectories to boost or penalize the rewards associated with a trajectory is proposed. Independent of imitation learning, trajectory classification and trajectory anomaly detection techniques~\cite{bian2019trajectory,wang2020anomalous} have been proposed in literature to determine if the path followed by a vehicle to travel between two locations conforms to the usual set of travel routes between those locations. In this paper, instead of determining confidences or recalculating rewards for demonstrated trajectories, we first partition demonstrated trajectories and then repair or discard trajectory parts based on a metric calculated from spatial and temporal features of trajectories. Some of the aforementioned techniques such as confidence measures, feasibility scores and trajectory anomaly metrics could also be used in conjunction with our technique to make the decision to repair or discard demonstrated trajectories.

Options or hierarchically abstract policies have been proposed as a framework to improve the planning quality and computation time of policies~\cite{sutton1999between}. Recently, the option-critic architecture~\cite{bacon2017option} has generalized the problem of determining options for a task using two components called the option and critic that work in tandem with each other. The option component evaluates the options based on current parameters while the critic component updates the parameters of the policy underlying options by calculating value and objective functions. In~\cite{daniel2016probabilistic} authors have proposed methods to automatically calculate options from data without using human-specified parameters and option-related information. Our work in this paper applies the framework of options to the address issues in adversarial reinforcement learning.

%The objective of imitation learning is to learn via supervised learning a process called training during which it explores the environment, observes the state-action-reward pairings received during the exploration, and finally selects the set of state-action-reward pairings that lead to higher expected rewards as its policy. 

\section{Imitation Learning with Adversarial Experts}
\label{sec:model}
{\bf Preliminaries.} We formalize the reinforcement learning framework using a Markov Decision Process (MDP) given by $(S, A, T, R, \gamma)$ where $S$ denotes the set of states and $A$ denotes the set of actions for the learning agent, $T$ denotes a state to action transition function specifying the forward dynamics model of the environment, where $T(s, a, s')$ is the probability of the agent reaching state $s'$ when it takes action $a$ at state $s$, $R: S \times A \rightarrow \Re$ denotes a reward function that gives a reward received by the agent by taking action $a$ at state $s$, and $\gamma$ is a discount factor. A policy $\pi: S \rightarrow [0,1]^{|A|}$ is a state to action mapping that prescribes a probability distribution $P(A)$ over the action set. The objective of the RL algorithm is to determine an optimal policy that maximizes the expected rewards, that is, $\pi^* = \arg \max_\pi \mathop{\mathbb{E}}(\sum_{t=0}^\infty \gamma^t R(s_t, a_t))$. Let  $P^* = P(s|\pi^*) = P(s, a^*=\pi^*)$ denote the probability distribution of state-action pairs while following the optimal policy $\pi^*$. In imitation learning, human experts provide demonstrations in the form of state-action sequences called trajectories that represent the policy. The $i$-th trajectory is denoted by $\tau_i^\pi = (s_{i,k},a_{i,k})_{k=0}^{H}$, where $\pi$ is the policy used to generate the actions in $\tau_i$ and $H$ denotes an episode's horizon or the average length of a trajectory. For the sake of legibility, in the rest of the paper, we use $a_{i,k} = \pi(s_{i,k})$ as a shorthand for $a_{i,k} = \arg \max_a \pi(s_{i,k})$.{\footnote{Usually the expert demonstrates actions, $a_{i,k}$, only and the states are given by the agent's forward dynamics model $T(s_{i,k}, a_{i,k}, s'_{i,k})$}} Let $\pi_\theta$ denote the policy learned using imitation learning where $\theta$ is a policy parameter (e.g., a set of weights in a policy network). The objective of imitation learning is to determine the optimal policy by finding an optimal policy parameter $\theta^*$ that minimizes the expected loss between the actions from the optimal policy provided via the expert demonstrations and the actions as per the learned policy $\pi_\theta$, that is, $\theta^* = \arg \min_\theta \mathbb{E}_{(s_{i,k}, a^*_{i,k}) \sim P^*} L(a^*_{i,k}, \pi_\theta(s_{i,k}))$. It is assumed that the expert performs its actions following the optimal policy, so $a^*_{i,k} = \pi^*(s_{i,k})$, and, consequently, the state-actions pairs in the expert trajectories conform to $P^*$, that is, $\forall i, k, (s_{i,k}, a^*_{i,k}) \sim P^*$. The value of policy $\pi$ is given by $V^\pi = \mathbb{E}[\sum_{i=0}^H R(s_i, a_i): a_i = \pi(s_i)]$.

For our problem setting, we consider a mix of benign and adversarial experts. Benign experts provide clean trajectories to the learner that follow the optimal policy and demonstrate the correct way to perform the task. We denote $\mathbb{T}_{clean}$ as the clean trajectory set, $\pi_{clean}$ as the policy learned via imitation learning from clean trajectories and $\tau_{clean}$ as a trajectory generated while using policy $\pi_{clean}$. An adversarial expert, on the other hand, demonstrates adversarial trajectories that are constructed by modifying clean trajectories, and, consequently, do not conform to the optimal or clean policy. The adversarial trajectory set, adversarial policy and an adversarial trajectory are denoted by $\mathbb{T}_{adv}$, $\pi_{adv}$ and, $\tau_{adv}$ respectively. By definition of $\pi_{adv}$ not being an optimal policy, it yields lower value than $\pi_{clean}$, that is, $\frac{V^{\pi_{adv}}}{V^{\pi_{clean}}} < 1$. %From the definition of an adversarial trajectory, $\forall \tau_i \in \mathbb{T}_{adv}$, $\exists a_{i,k}:  a_{i,k} \neq \pi_{clean}(s_{i,k})$, and $(s_{i,k}, a^*_{i,k}) \sim P^*$ does not hold any more. 
For our problem, we denote a trajectory set as: $\mathbb{T} = (\mathbb{T}_{clean} \cup \mathbb{T}_{adv}, \eta, \{\gamma_i\})$, where $\eta \in [0, 1]$ denotes the fraction of trajectories that have been modified and $\gamma_i\in[0, 1]$ denotes the fraction within the $i$-th trajectory that has been modified. Mathematically, $\eta = \frac{|\mathbb{T}_{adv}|}{|\mathbb{T}_{clean} \cup \mathbb{T}_{adv}|}$ and $\gamma_i = \frac{\sum k}{|\tau_i|}: (s_{i,k}, a_{i,k}) \in \tau_i \wedge a_{i,k} \neq \pi_{clean}(s_{i,k})$. The values of $\mathbb{T}_{clean} \cup \mathbb{T}_{adv}, \eta$ and $\gamma_i$ are known to the adversarial expert while the learning agent only knows $\mathbb{T}_{clean} \cup \mathbb{T}_{adv}$.

We have divided our proposed technique into two parts. First, we describe the options framework to learn policies for sub-tasks from partial trajectories. Then, we develop a trajectory divergence measure between a demonstrated trajectory and known or clean trajectories that can be used to decide whether to accept or reject the demonstrated trajectory parts. 

\subsection{Policy Repair Using Options}
We propose an options-based framework for policy repair where, instead of learning a policy over the entire state-action space, the state-action space is partitioned into subsets and a policy is learned for each part. Without loss of generality, we assume that the partition is done temporally - a trajectory $\tau$ is partitioned into $M$ equal parts, and the $i$-th part ($i=0, ..., M-1$) is denoted by $\tau_i$. Intuitively, this partition corresponds to dividing the end-to-end or full-horizon task into subtasks. The main idea in options is to learn a policy for each sub-task. Formally, an option for the $i$-th part is defined as $\omega_i = (I_i, \pi_i, \beta_i)$, where  $I_i$ is the set of initiation or start states for sub-task $i$, $\pi_i$ is the optimal policy for solving sub-task $i$, and $\beta_i$ is the termination or end states for sub-task $i$. As before, $\pi_{\theta_i}^*$ is learned via imitation learning and given by $\theta_i^* = \arg \min_{\theta_i} \mathbb{E}_{s, a^* \sim P_i^*(s)} (L(a^*, \pi_{\theta_i}(s))$ where $P_i^*(s) = P(s|\pi_i^*)$. 

\subsubsection{Using Trajectory Divergence to Accept/Reject Trajectories}
A core aspect of our options-based policy repair technique is to be able to determine the divergence between an unknown (whether it is benign or adversarial) demonstrated trajectory and a clean trajectory, that is, one that is guaranteed to be non-adversarial. This divergence measure can then be used to decide whether to accept or reject the demonstrated trajectory. However, a straightforward approach of making the trajectory accept/reject decision based on a single metric-based divergence measure might not work. For instance, an adversarial expert might demonstrate trajectories that have low divergence with clean trajectories, but inject a few incorrect moves or actions at key states in the trajectories that result in the agent either failing to do the task or doing it sub-optimally. 
Again, a demonstrated trajectory might represent a previously unseen but correct and possibly improved way of doing the task. This trajectory would have a higher divergence measure with known, clean trajectories and if the accept/reject decision is based on the divergence measure only, it would end up getting an incorrect, reject decision. To address these challenges, we propose a divergence measure that combines two commonly used trajectory divergence measures with a supervised learning-based classification technique, as described below. 

\begin{algorithm}[thb!]
\SetAlgoLined
\SetKwInOut{Input}{input}\SetKwInOut{Output}{output}
\SetKwProg{myproc}{Procedure}{}{}
\Input{$T_{clean}, T_{demo}$: Clean and expert trajectory  sets} 
\Output{$\Omega$: set of options}
\BlankLine
\SetKwFunction{proc}{Repair-options}{}
\myproc{\proc{$T_{clean}, T_{demo}$}}{
$\Omega, D_{chain} \leftarrow \{\emptyset\}$\\
%Sample $T_{clean}$ and $T_{demo}$ from $\mathbb{T}_{clean}, \mathbb{T}_{demo}$ resp.\\
Split each $\tau_{clean} \in T_{clean}$ and $\tau_{demo} \in T_{demo}$ into $M$ equal parts\\
\For{ $i= 0 \ldots M-1$}
{
    $OC_i \leftarrow OC(\tau_{clean, i}, \tau_{demo, i})$\\
    $FD_i \leftarrow FD(\tau_{clean, i}, \tau_{demo, i})$\\
    \If{($\chi(OC_i, FD_i)$ == {\tt Accept}) OR ($\chi(OC_i, FD_i)$ == {\tt Reject} AND $\frac{R_{\tau_{demo}}}{R_{\tau_{clean}}} > 1 - \epsilon_p$)} {
        
            $T_{train, i} \leftarrow T_{clean,i} \cup  T_{demo,i}$\\
            $\pi_i \leftarrow$ Train sub-policy $i$ using $T_{train, i}$ via imitation learning\\ 
            $I_i, \beta_i \leftarrow$ Initiation and terminating states from $T_{train, i}$\\
            $\omega_i \leftarrow (I_i, \pi_i, \beta_i)$\\
            $\Omega \leftarrow \Omega \cup \omega_i$
        
    }
}
\For{ $i= 1 \ldots M-1$}{
    \For {every $s_i \in \beta_i$}
    {        
        $s^*_j \leftarrow \arg \min_{s_j \in  I_{i+1}} ||s_i - s_j||$\\
        $D_{chain} \leftarrow D_{chain} \cup (s_i, s^*_j)$
    }
}
return $\Omega, D_{chain}$
}
\caption{Trajectory repair using options}
\label{algo:traj_repair}
\end{algorithm}

{\bf Occupancy Measure (OC).} The first trajectory divergence measure we use is the {\bf occupancy measure}~\cite{puterman2014markov}. It represents the number of times state-action pairs along a given expert trajectory are visited while using the (clean) policy. The occupancy measure of a demonstrated trajectory $\tau = ((s_0, a_0) (s_1, a_1), \ldots, (s_{|\tau|}, a_{|\tau|}))$ with respect to a clean trajectory $\tau_{clean}$ generated while following the clean policy $\pi_{clean}$, is given by : \[OC_\tau = \sum_{(s_i, a_i) \in \tau_{clean}} \pi^*(a_i|s_i) \sum_{t=0}^{|\tau|} \gamma^t p(s_t=s_i | \pi_{clean}),\] where $\gamma \in [0, 1]$ is a discount factor. Clearly, $OC_\tau$ has higher values when the demonstrated trajectory, $\tau$, is closer or similar to the clean trajectory, $\tau_{clean}$. The minimum value of $OC_\tau = 0$ happens when there is no overlap between the state-action pairs of the two trajectories. The occupancy measure is a suitable metric for making the accept/reject decision of a demonstrated trajectory if it overlaps with many state-action pairs of clean trajectories. However, a limitation of using it as the only decision variable is that if the demonstrated trajectory is non-adversarial and similar to a clean trajectory, but overlaps with very few or no state-action pairs in it, the occupancy measure would be close to or equal to zero and give an incorrect decision of rejecting the trajectory. 
    
{\bf Fr\'echet Distance (FD).} Our second trajectory divergence measure is the {\bf Fr\'echet distance}~\cite{alt1995computing}. It gives the distance between two polylines while considering the spatial and temporal ordering of the points on them. Mathematically, the Fr\'echet distance between an expert trajectory $\tau$ and a clean trajectory $\tau_{clean}$ is given by:
\[FD_{\tau} = \min_{\alpha, \beta} \max_{t \in [0,1]} d(\tau(\alpha(t)), \tau_{clean}(\beta(t))),\] 
where, $d(\cdot)$ gives the Euclidean distance or L2 norm between two trajectory points on $\tau$ and $\tau_{clean}$ respectively. $\alpha, \beta$ are functions that take an argument $t \in [0,1]$ and return an index into $\tau$ and $\tau_{clean}$ respectively, with $\alpha(0) = \beta(0) = 0$, and $\alpha(1) = |\tau|, \beta(1) = |\tau_{clean}|$. The Fr\'echet distance calculation iterates over different functions for $\alpha$ and $\beta$, determines the maximum distance between ordered pairs of points on $\tau$ and $\tau_{clean}$ for each $\alpha$ and $\beta$ combination iterated over, and, finally, returns the minimum of these maximum distances. When both expert and clean trajectories are identical, the Fr\'echet distance has its smallest value, $0$. As the two trajectories get further apart, the Fr\'echet distance increases. For the last example from the previous paragraph, using the Fr\'echet distance rectifies the incorrect decision given by occupancy measure as the Fr\'echet distance for a demonstrated trajectory with high similarity but little or no overlap in state-action pairs with a clean trajectory would have a low value and yield a correct decision to accept the trajectory. 

To make an accept/reject decision of a trajectory based on its occupancy measure and Fr\'echet distance values, we train a classifier, $\chi: OC \times FD \rightarrow \{{\tt Accept, Reject}\}$ via supervised learning. The classifier's training set contains the $OC$ and $FD$ values sampled from different clean and adversarial trajectories, along with a label, $\lambda_\tau$ for each trajectory sample, given by:
\begin{equation}
    \lambda_\tau =  \begin{cases}
        \text{{\tt Accept}} & \text{if } R(\tau) \geq (1-\epsilon_p)R_{max}\\
        \text{{\tt Reject}} & \text{otherwise} \nonumber
\end{cases}
\end{equation}  

{\bf Handling Benign Divergent Trajectories.} The classifier $\chi$ suffices to admit trajectories based on the similarity of their spatio-temporal features to known, benign trajectories. However, a demonstrated trajectory that shows a novel way to perform the task and is suitable for learning from, might have a high divergence measure and, consequently, get rejected by the classifier. To address these false positives, we augment the classifier's prediction with a special condition that reverses only the reject decisions on a trajectory if the ratio of the returns (sum of rewards) between the demonstrated and clean trajectories is above a fraction $1-\epsilon_p$. The advantage of using the return ratio only is that it can be calculated quickly using the agent's reward function and demonstrated trajectory data, without requiring access to the agent's policy or value functions that require complex, time-consuming calculations.

Algorithm~\ref{algo:traj_repair} gives the pseudo-code algorithm for repairing trajectories with our options-based framework using the above divergence measures and trajectory accept/reject decision classifier. Given a set of guaranteed, clean trajectories, $T_{clean}$ and a set of demonstrated trajectories we first split trajectories from each set into $M$ parts (line $3$). For each part, we determine if it can be accepted into the training set for the imitation learning algorithm using the classifier's 'Accept' prediction or return ratio criteria (lines $5-7$). If acceptable, the demonstrated trajectories are included with the clean trajectories for training the policy $\pi^*_i$  for sub-task $i$ via imitation learning (line $8-9$). The initiation and termination states for option $i$ are also recorded along with policy $\pi^*_i$ within option $\omega_i$. An important requirement for using options is option chaining which determines when to terminate option $\omega_i$ and how to select the next option $\omega_{i+1}$, so that an end-to-end policy can be formed in the state-action space of the problem. While chaining is done at policy execution time~\cite{KonidarisKGB12}, we create a dictionary $D_{chain}: S \rightarrow S$ while creating the set of options to speed up execution. $D_{chain}$ is constructed in Lines $15-18$ in Algorithm~\ref{algo:traj_repair} by recording the closest state $s_j \in I_{i+1}$ is closest in terms of L2 norm distance to a state $s_i \in \beta_i$. 

\subsubsection{Option Chaining}
 
\begin{algorithm}[tbh]
\SetAlgoLined
\SetKwInOut{Input}{input}\SetKwInOut{Output}{output}
\SetKwProg{myproc}{Procedure}{}{}
\Input{$s_0, g, \Omega, D_{chain}$: Start state, goal state, option set, option chain dictionary} 
\Output{{\tt Task-Success} or {\tt Task-Failure}}
\BlankLine
\SetKwFunction{proc}{Chain-options}{}
\myproc{\proc{$s_0, g, \Omega, D_{chain}$}}{
$t, i \leftarrow 0$\\
$\pi_{cur} \leftarrow \pi_i : (I_i, \pi_i, \beta_i) \in \Omega$\\
$s_{cur} \leftarrow s_0$\\
\Do{$||s_{cur} - g|| \geq \epsilon_{chain} \wedge t \leq T_{max}$}{
    $a_{cur} \leftarrow \arg \max_a \pi_{cur}(s_{cur})$\\
    $s_{cur} \leftarrow \arg \max_{s'} T(s_{cur}, a_{cur}, s')$\\
    $S_{end} \leftarrow \{s_i: ||s_{cur} - s_i|| \leq \epsilon_{chain} \wedge s_i \in \beta_i$\}\\    
    \If{$S_{end} \neq \{\emptyset$\}}{
        $s_{end} \leftarrow \arg \min_{s_i \in S_{end}} ||s_{cur} - s_i||$\\
        $s_{cur} \leftarrow D_{chain}(s_{end})$\\
        $\pi_{cur} \leftarrow \pi_{i+1}$\\
    }
}
\If{$t > T_{max}$}{
    return {\tt Task-Failure}
}
return {\tt Task-Success}
}
\caption{Chaining Options at run-time}
\label{algo:chain-options}
\end{algorithm}

Algorithm~\ref{algo:chain-options} shows the option chaining at run-time to enable executing successive policies for sub-tasks using options. As shown in Lines $8-12$ of Algorithm~\ref{algo:chain-options}, to determine if policy $\pi_i$ in option $\omega_i$ is about to terminate, a state $s_{cur}$ that is reached by the agent while executing $\pi_i$  is checked for proximity within an L2 norm distance of $\epsilon_{chain}$ from any state in the termination set $\beta_i$. If any such states exist in $\beta_i$, the closest such state to $s_{cur}$, $s_{end}$, is selected (line $10$) and $s_{cur}$ is updated to a state in the initiation set of the next option $\omega_{i+1}$ given by $D_{chain}(s_{end})$ (line $11$). The current option is also updated to the option for the next sub-task (line $12$).

\section{Theoretical Analysis}
\label{sec:theoritical_analysis}
In this section, we formalize our trajectory repair technique described in Section~\ref{sec:model}. First, we show that using trajectory divergence measure alone gives a weak condition for the accept/reject decision for demonstrated trajectories. We then show that augmenting this decision with a rewards-based rule (following Algorithm~\ref{algo:traj_repair}, Line 7) guarantees that accept/reject decisions are consistent with benign and adversarial trajectories. Finally, we show that the above results remain valid for part trajectories so that they can be applied to our options-based, trajectory repair technique.

{\bf Definition \arabic{defcounter}.} {\em Dominated Policy.} Given two policies $\pi$ and $\pi'$ we say $\pi$ is dominated by $\pi'$ if $\frac{V^\pi}{V^{\pi'}} < 1-\epsilon_{p}$, where $\epsilon_{p}$ is a constant. We denote this in shorthand as $\pi \prec \pi'$.
\stepcounter{defcounter}

{\bf Definition \arabic{defcounter}.} {\em Divergent Trajectories.} Let $\tau^\pi$ and $\tau^{\pi'}$ represent two trajectories that are sampled from two policies $\pi$ and $\pi'$. We say $\tau^\pi$ and $\tau^{\pi'}$  are {\em divergent} if $D(\tau^\pi, \tau^{\pi'}) > \delta$, where $D$ is a divergence measure between $\tau^\pi$ and $\tau^{\pi'}$ and $\delta$ is a constant.
\stepcounter{defcounter}

{\bf Definition \arabic{defcounter}.} {\em Local Policy Repair Function.} Given state $s$ and two policies $\pi$ and $\pi'$ with $\pi \prec \pi'$, a local policy repair function is a transformation $f_{rep}: S \times [0, 1]^{|A|} \rightarrow [0, 1]^{|A|}$, such that, $\tilde{D}(\pi_s || f_{rep}(s, \pi'_s)) < \epsilon_{div}$, where $\tilde{D}$ is a distance measure between two probability distributions.{\footnote{Note that $f_{rep}(s, \pi'_s)$ transforms $\pi'$ to a new policy, say $\pi''$}}
\stepcounter{defcounter}

{\bf Definition \arabic{defcounter}.} {\em $\epsilon$-repair set:} Given an initial policy $\pi$ and a target policy $\pi^{tar}$, the {\em $\epsilon$-repair set} for $\pi, \pi^{tar}$, $\rho_{\pi \rightarrow \pi^{tar}}$, is a set of states such that the policy $\pi'$ obtained by applying $f_{rep}(s, \pi)$ to every $s \in \rho_{\pi \rightarrow \pi^{tar}}$ satisfies $\frac{V^{\pi'}}{V^{\pi^{tar}}} \geq 1-\epsilon_{p}$.
\stepcounter{defcounter}

%A minimal $\epsilon$-repair set is an $\epsilon$-repair set with minimum cardinality.

{\bf Theorem \arabic{thmcounter}.} If $\pi \prec \pi'$, then trajectories  $\tau^\pi$ and $\tau^{\pi'}$ sampled from $\pi$ and $\pi'$ respectively are divergent .\\
\stepcounter{thmcounter}
{\em Proof.} (By contradiction.)
Let us suppose $\pi \prec \pi'$, but trajectories $\tau^\pi$ and $\tau^{\pi'}$ are not divergent, that is, $D(\tau^\pi, \tau^{\pi'}) \leq \delta$. Without loss of generality, we assume $\delta = 0$. This implies that the divergence measure between $\tau^\pi$ and $\tau^{\pi'}$ is zero, and, consequently, $\forall i, \, s_i^{\tau^\pi} = s_i^{\tau^{\pi'}}$. Now, from the definition of a dominated policy in Definition $1$, it follows that $V^\pi \neq V^{\pi'}$. Recall, $V^\pi= \mathbb{E}[\sum_{i=0}^H R(s_i, a_i): a_i = \pi(s_i)]$, and, so, there must be at least one time-step, $i$, at which, $R(s_i^\pi, a_i^\pi) \neq R(s_i^{\pi'}, a_i^{\pi'})$. This implies, either $s_i^\pi \neq s_i^{\pi'}$, or, $s_i^\pi = s_i^{\pi'}$, but $a_i^\pi \neq a_i^{\pi'}$. The latter case, implies different actions are taken at state $s_i$ by policies $\pi$ and $\pi'$, which leads to different next states $s_{i+i}^\pi \neq s_{i+1}^{\pi'}$, reached by policies $\pi$ and $\pi'$. In both cases, there are at least two states on trajectories generated from $\pi$ and $\pi'$ that are distinct from each other, that is, $s_i^{\tau^\pi} \neq s_i^{\tau^{\pi'}}$, for at least some $i$. This contradicts our assumption, $\forall i, \, s_i^{\tau^\pi} = s_i^{\tau^{\pi'}}$. Hence proved. $\square$

However, we note that converse of Theorem $1$ is not valid - when $D(\tau^\pi, \tau^{\pi'}) > \delta$, it is not guaranteed that policy $\pi'$ will dominate $\pi$. We give an informal proof sketch: if $D(\tau^\pi, \tau^{\pi'}) > \delta$, there must be at least one $i$ where $s_i^{\tau^\pi} \neq s_i^{\tau^{\pi'}}$. We cannot make any guarantees about the relative rewards at these states while using policies $\pi'$ and $\pi$. If $R(s_i^\pi, a_i^\pi) > R(s_j^{\pi'}, a_j^{\pi'}), s_i \neq s_j$, we could get $V^\pi > V^{\pi'}$, which would imply that $\pi'$ is dominated by $\pi$. On the other hand, if $R(s_i^\pi, a_i^\pi) < R(s_j^{\pi'}, a_j^{\pi'})$, $\pi'$ dominates $\pi$. This means that trajectory divergence is a necessary, but not a sufficient condition for policy dominance. This necessitates an additional condition to select states from $S$ to construct the $\epsilon$-repair set. For this, we propose the following rule: 

{\bf Rule $1$.} For a state $s_i^\pi$ to be added to $\epsilon$-repair set, $\rho_{\pi \rightarrow \pi'}$, $\frac{R(s_i^\pi,\pi_{s_i^\pi})}{R(s_i^{\pi'}, \pi_{s_i^{\pi'}})} < 1-\epsilon_p$.

The above rule states that a state can be added to the $\epsilon$-repair set if the reward at that state by selecting an action using policy $\pi$ is lower than selecting an action using policy $\pi'$. Based on this rule, we have the following theorem about the convergence of trajectories based on their divergence measure.

Let $M_{max}$ denote the maximum number of states in $S$ where $R(s_i^\pi,\pi_{s_i^\pi}) < R(s_i^{\pi'}, \pi_{s_i^{\pi'}})$.

{\bf Theorem \arabic{thmcounter}.} If Rule $1$ is applied $M$ times to build the $\epsilon$-repair set $\rho_{\pi \rightarrow \pi'}$, then as $M \rightarrow M_{max}$, $\frac{V^\pi}{V^{\pi'}} \rightarrow 1$ and $D(\tau^\pi, \tau^{\pi'}) \rightarrow 0$.\\
\stepcounter{thmcounter}

{\em Proof.}{\footnote{For legibility, we give the proof for $\epsilon_p = 0$, it can be extended easily to $\epsilon_p = 0^+$.}} Recall that $V^\pi= \mathbb{E}[\sum R(s_i^\pi, \pi_{s_i^\pi})]$ and $V^{\pi'}= \mathbb{E}[\sum R(s_i^{\pi'}, \pi'_{s_i^{\pi'}})]$. The difference between these two terms can be written as:

\begin{align*}
V^{\pi'} - V^\pi  = & \mathbb{E}[\sum R(s_i^{\pi'}, \pi'_{s_i^{\pi'}}) - R(s_i^\pi, \pi_{s_i^\pi})] \\
= & \mathbb{E}[R(s_1^{\pi'}, \pi_{s_1^{\pi'}}) + ... + R(s_k^{\pi'}, \pi_{s_k^{\pi'}})  + ... + R(s_M^{\pi'}, \pi_{s_M^{\pi'}})] \\
& - \mathbb{E}[R(s_1^\pi, \pi_{s_1^\pi}) + ... + R(s_k^\pi, \pi_{s_k^\pi})  + ... + R(s_M^\pi, \pi_{s_M^\pi})] 
\end{align*}

We use $\Delta V_0^{\pi' - \pi}$ as a shorthand to denote the initial value of $V^{\pi'} - V^\pi$ (before applying Rule $1$), and, $\Delta V_1^{\pi' - \pi}$ as its value after applying Rule $1$ once, $\Delta V_2^{\pi' - \pi}$ as its value after applying Rule $1$ twice, and, so on. If we select states $(s_k^\pi, s_k^{\pi'})$ via Rule $1$ and apply $f_{rep}(s_k^\pi, \pi_{s_k^\pi})$, then because $ R(s_k^{\pi'}, \pi_{s_k^{\pi'}}) - R(s_k^\pi,\pi_{s_k^\pi}) > 0$, therefore, $\Delta V_0^{\pi' - \pi} > \Delta V_1^{\pi' - \pi}$. Similarly,  $\Delta V_1^{\pi' - \pi} > \Delta V_2^{\pi' - \pi}$. If we continue in this manner, $V^{\pi'} - V^\pi$ becomes successively smaller and smaller. Finally, when $f_{rep}()$ has been applied at most $M_{max}$ times, $\Delta V_{M_{max}}^{\pi' - \pi}= 0$. At this point, $V^{\pi'} = V^\pi$, or $\frac{V^\pi}{V^{\pi'}} = 1$. In the limiting case, when state pairs $(s_k^\pi, s_k^{\pi'})$ have $ R(s_k^{\pi'}, \pi_{s_k^{\pi'}}) - R(s_k^\pi,\pi_{s_k^\pi}) \approx 0^+$, we get, $\frac{V^\pi}{V^{\pi'}} \rightarrow 1$.

In a similar manner, applying $f_{rep}(s_k^\pi, \pi_{s_k^\pi})$ makes $\pi_{s_k} = \pi'_{s_k}$, and, consequently, the same state is reached by taking action $\pi'_{s_k}$ at $s_k^\pi$. This makes, $D_0(\tau^\pi, \tau^{\pi'}) > D_1(\tau^\pi, \tau^{\pi'}) > D_2(\tau^\pi, \tau^{\pi'}) > ...$, where the subscript denotes the number of times Rule $1$ and $f_{rep}()$ have been applied. When $f_{rep}()$ has been applied $M_{max}$ times, we get $D_{M_{max}}(\tau^\pi, \tau^{\pi'}) = 0$, and, in the limiting case, when state pairs $(s_k^\pi, s_k^{\pi'})$ have $ R(s_k^{\pi'}, \pi_{s_k^{\pi'}}) - R(s_k^\pi,\pi_{s_k^\pi}) \approx 0^+$, $D(\tau^\pi, \tau^{\pi'}) \rightarrow 0$. $\square$.

{\bf Lemma \arabic{thmcounter}.} If policies $\pi$ and $\pi'$, $\pi \prec \pi'$, are divided into sub-policies $\pi_1, ~\pi_2, ... \pi_M$ and $\pi'_1, \pi'_2 ... \pi'_M$, then for at least one interval $m \in \{1, ..., M\}$, $\pi_m \prec \pi_m'$
\stepcounter{thmcounter}

{\em Proof.} (by contradiction) From Definition $1$, if $\pi \prec \pi'$, then $V^\pi < V^{\pi'}$.{\footnote{For simplicity and without loss of generality, we slightly relax Definition $1$ by assuming $\epsilon_d=0$, which gives $\frac{V^\pi}{V^{\pi'}}<1$} Suppose $\pi \prec \pi'$ and policies $\pi$ and $\pi'$ are divided into sub-policies, $\pi_1, \pi_2$ and $\pi'_1, \pi'_2$ respectively, and, both sub-policies of $\pi$ are not dominated. That is, $V^\pi_1 \geq V^{\pi'}_1$ and $V^\pi_2 \geq V^{\pi'}_2$. Rearranging and adding terms of the last two inequalities, we get, $V^\pi_1 - V^{\pi'_1} + V^\pi_2 - V^{\pi'}_2 \geq 0$, or, $(V^\pi_1 + V^\pi_2) - (V^{\pi'_1} + V^{\pi'}_2) \geq 0$. Substituting, $(V^\pi_1 + V^\pi_2) = V_\pi$ and $V^{\pi'_1} + V^{\pi'}_2 = V^{\pi'}$, we get $ V^\pi -  V^{\pi'} > 0$, or, $ V^\pi >  V^{\pi'}$, which contradicts the definition of $\pi \prec\pi'$. Therefore, our assumption that $V^\pi_1 \geq V^{\pi'}_1$ and $V^\pi_2 \geq V^{\pi'}_2$ (both sub-policies are not dominated) is incorrect, and at least one of the sub-policies must be dominated. This proof can be easily extended beyond two sub-policies by induction. $\square$

\omitit{
{\bf Theorem \arabic{thmcounter}.} Repairing partial policy via options is faster than repairing full horizon policy.\\
\stepcounter{thmcounter}
{\em Proof.}\\
}
\section{Experimental Results}
\label{sec:expts}

\subsection{Experimental Setup}

{\bf Environment.} For evaluating our option-based adversarial RL algorithm, we used the \texttt{LunarLander-v2} environment available within AI Gym~\cite{brockman2016openai}. The problem consists of landing an airborne two-legged spacecraft at a specific location called the landing pad within a 2D environment akin to the surface of the Moon. The state space consists of an $8$-dimension vector given by the 2-D coordinates of the center of the spacecraft, 2-D linear velocity, orientation and angular velocity and whether both legs of the spacecraft are on the ground. The initial state of the spacecraft consists of random coordinates towards the top of the environment and random initial velocity. The action space of the spacecraft consists of four actions: to fire its main, left or right engines or do nothing (no-op). The agent receives a reward of $320$ points of landing on both legs on the landing pad, a penalty of $-100$ points for crashing, while maneuvering the spacecraft incurs a penalty of $-0.3$ for using the main engine and $-0.03$ for the left or right engine. For our baseline reinforcement learning algorithm we used the deep Q-network learning (DQN) algorithm available via stable baselines~\cite{stable-baselines}. The algorithms were implemented using the following open source libraries: Tensorflow $1.15$, OpenAI Gym $0.18$ and Stable Baselines $2.10$.

{\bf Generating clean and adversarial trajectories.} To generate clean trajectories we trained a Deep Q-network (DQN) algorithm in the LunarLander environment for $2.5 \times 10^5$ time-steps, all other algorithm hyper-parameters were set to the values given in RL Baselines Zoo~\cite{RLBaselineszoo}. We generated $1000$ clean trajectories. These trajectories were then modified using the adversarial trajectory modification algorithms described in Section~\ref{sec:adv_traj_attacks}. For the adversarial attacks, we used $\eta = \{0.3, 0.6, 0.9\}$, $\gamma_i = \{0.3, 0.6, 0.9\}$, and attack location as $\{$BEG, MID, END, FLP$\}$ giving rise to $36$ different adversarial trajectory sets, each comprising $1000$ trajectories.  
We then trained policies via imitation learning with these adversarial trajectory sets.

\subsubsection{Trajectory Modification Attacks by Adversary}
\label{sec:adv_traj_attacks}
We considered two adversarial attack strategies for modifying expert demonstrations: 1) A directed attack strategy that requires access only to clean trajectories demonstrated by a benign expert, 2) A gradient-based attack strategy that requires access to the learner's policy network and rewards.
%Given a clean trajectory set $\mathbb{T}_{clean}$, an adversary of either type first selects $\eta |\mathbb{T}_{clean}|$ trajectories to perturb by applying a perturbation function. 

\begin{algorithm}[ht!]
\SetAlgoLined
\SetKwInOut{Input}{input}\SetKwInOut{Output}{output}
\SetKwProg{myproc}{Procedure}{}{}
\Input{$\mathbb{T}_{clean}$: clean trajectory set, \\
$\eta$: Fraction of traj. set to be modified, \\
$\gamma_i$: Fraction of location in each trajectory $\tau_i$ to be modified\\
$att_{loc}: \{begin, center, end\}$: attack start location}
\Output{$\mathbb{T}_{adv}$: Adversarial trajectory set}
\BlankLine
\SetKwFunction{proc}{adv-traj-directed}{}
\myproc{\proc{$\mathbb{T}_{clean}, \eta, \gamma_i$}}{
$\mathbb{T}_{adv} \leftarrow \emptyset$\\
$\{\tau_i\} \leftarrow$ Select $\eta|\mathbb{T}_{clean}|$ trajectories from $\mathbb{T}_{clean}$\\
\For {each $\tau_i$} {
	$K \leftarrow$ Select $\gamma_i|\tau_i|$ locations (indices) using $att_{loc}$\\
	\For {each $k$ in $K$} {	 
		$\tau'_i \leftarrow \tau_i: a'_{i,k} \leftarrow \phi(a_{i,k})$ 
		}
	Add $\tau'_i$ to $\mathbb{T}_{adv}$\\
	}
return $\mathbb{T}_{adv}$
}
\caption{Adversarial directed trajectory modification.}
\label{algo_directed_traj}
\end{algorithm}

\noindent
A {\bf directed attack} targets sequential locations inside a trajectory starting from an attack start location, that could be either at the beginning (BEG), middle (MID) or end (END) of the trajectory. The actions at $\gamma_i|\tau_i|$ consecutive locations following the attack start location are then modified using an action modification function $\phi: A \rightarrow A$, given by  $\phi(a_{i,k}) = \arg \max_{s'} \Delta_s(s_{i,k}, s')$ where, $s' = \arg \max_{a'} T(s_{i,k}, a', s_{i,k+1)})$. That is, $\phi$ replaces $a_{i,k}$ with the action that takes the agent to a state $s'$ that is farthest along a state-distance metric, $\Delta_s$, from $s_{i,k+1}$. The directed attack is a straightforward, fast, yet effective attack as it does not require the adversary to have information about the learner's reward function. While it can be realized by the adversary with a lower attack budget it can also be detected relatively easily by the learner.
%Note that the transformations have to be done in a sequential manner to ensure that the states reached by taking the modified actions are still valid according to the forward dynamics model of the environment. {\footnote{In effect, action perturbation modifies actions along with associated states along the modified action path. We do not consider state-only perturbations as modifying only states, but not actions would violate the forward dynamics model of the environment.}}
\begin{algorithm}[ht!]
\SetAlgoLined
\SetKwInOut{Input}{input}\SetKwInOut{Output}{output}
\SetKwProg{myproc}{Procedure}{}{}
\Input{$\mathbb{T}_{clean}$: clean trajectory set, \\
$\eta$: Fraction of traj. set to be modified, \\
$\gamma_i$: Fraction of location in each trajectory $i$ to be modified}
\Output{$\mathbb{T}_{adv}$: Adversarial trajectory set}
\BlankLine
\SetKwFunction{proc}{adv-traj-gradients}{}
\myproc{\proc{$\mathbb{T}_{clean}, \eta, \gamma_i$}}{
$\mathbb{T}_{adv} \leftarrow \emptyset$\\
$\{\tau_i\} \leftarrow$ Select $\eta|\mathbb{T}_{clean}|$ trajectories from $\mathbb{T}_{clean}$\\
\For {each $\tau_i$}{
	$grads \leftarrow \frac{\partial r_j}{\partial o_j},\quad j=1...|\tau_i|$\\
	\For {$\gamma_i$ iterations} {
		$j_{min}, j_{max} \leftarrow \arg min(grads), \arg max(grads)$\\
		swap($a_{j_{min}}, a_{j_{max}}$) in $\tau_i$
	}
	Add $\tau_i$ (after swaps) to $\mathbb{T}_{adv}$\\
}
return $\mathbb{T}_{adv}$
}
\caption{Adversarial gradient-based trajectory generation.}
\label{algo_grad_traj}
\end{algorithm}

\noindent 
{\bf Gradient-based Attack.} The gradient-based technique is inspired the hot-flip (FLP) technique~\cite{ebrahimi-etal-2018-hotflip} for text perturbation. The technique identifies the minimum number of characters and their locations within a text string that need to be modified so that the text string gets mis-classified by a supervised learning-based model. We apply a similar idea for our gradient-based attack where the adversary identifies locations or indices in the trajectory that need to be modified, by considering the gradient of the objective or reward function with respect to the observation, denoted by $\frac{\partial r_{i,k}}{\partial o_{i,k}}, k = 1...|\tau_i|$, and swap the observations corresponding to maximum and minimum gradients for $\gamma_i|\tau_i|$ iterations. The pseudo-code for the gradient-based attack is shown in Algorithm~\ref{algo_grad_traj}, 

Note that for the gradient-based attack, the adversary needs to have knowledge about the learner's reward function. We note that researchers have proposed sophisticated but also more computationally complex attacks for modifying actions~\cite{lee2020spatiotemporally, lin2017tactics} that aim to reduce the reward received by an RL agent, although not within the context of imitation learning. Our attacks are computationally simpler but still achieve the desired effect of reducing the learner's rewards. The policy repair technique proposed in the paper could also be used in conjunction with any of these attacks.

%for adversarial text generation: successful adversarial text perturbations can be generated by swapping characters that correspond to the maximum and minimum gradient of the objective or loss function with respect to the input within a string of text. For our RL context, the input to the RL policy is an observation, while the objective is to determine the policy parameters that maximize the expected reward. Therefore,

%Benign and adversarial expert trajectories are drawn from corresponding expert polices, $\pi_E^{ben}$ and $\pi_E^{adv}$, that are not known by the learning agent. The objective of the learning agent is to use the expert trajectories to calculate a learner policy, $\pi_L$, for performing the task.

Note that because the adversary modifies clean trajectories to generate adversarial trajectories, it knows the partition of $\mathbb{T}$ into $\mathbb{T}_{clean}$ and $\mathbb{T}_{adv}$ and can calculate $\gamma_i$ and $\eta$. The learner on the other hand does not know the partition and is not aware of these parameters. 

\omitit{
\begin{figure}[htb!]
\begin{center}
\begin{tabular}{c}
\hspace*{-1.2in}
\includegraphics[width=7.5in]{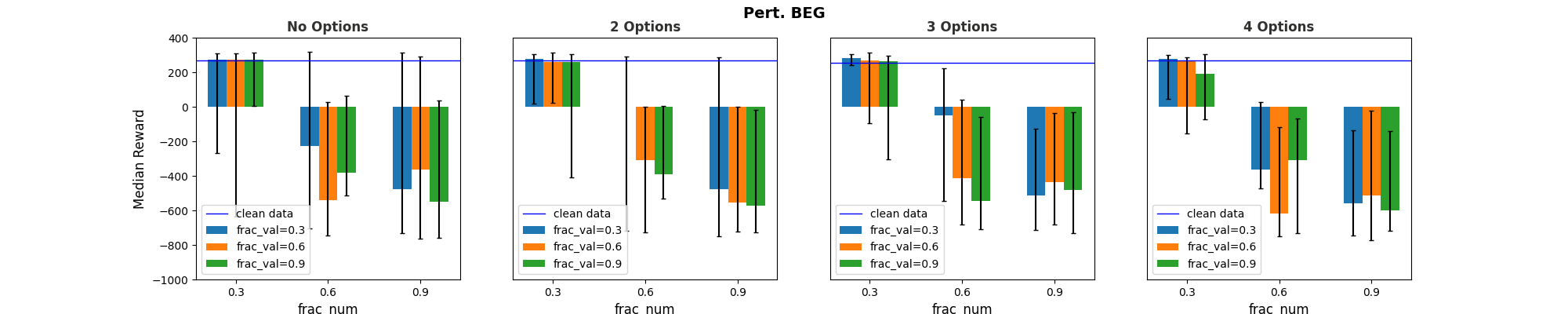}\\
\hspace*{-1.2in}
\includegraphics[width=7.5in]{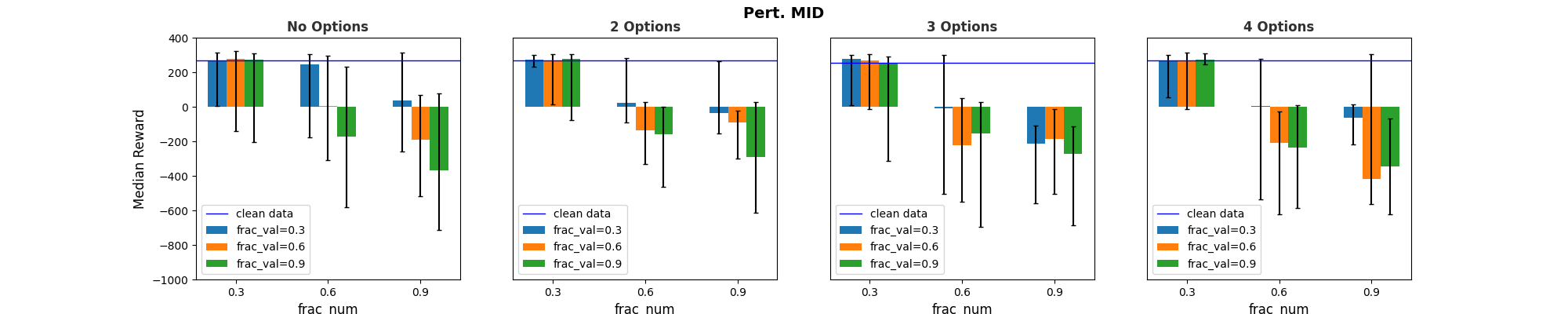}\\
\hspace*{-1.2in}
\includegraphics[width=7.5in]{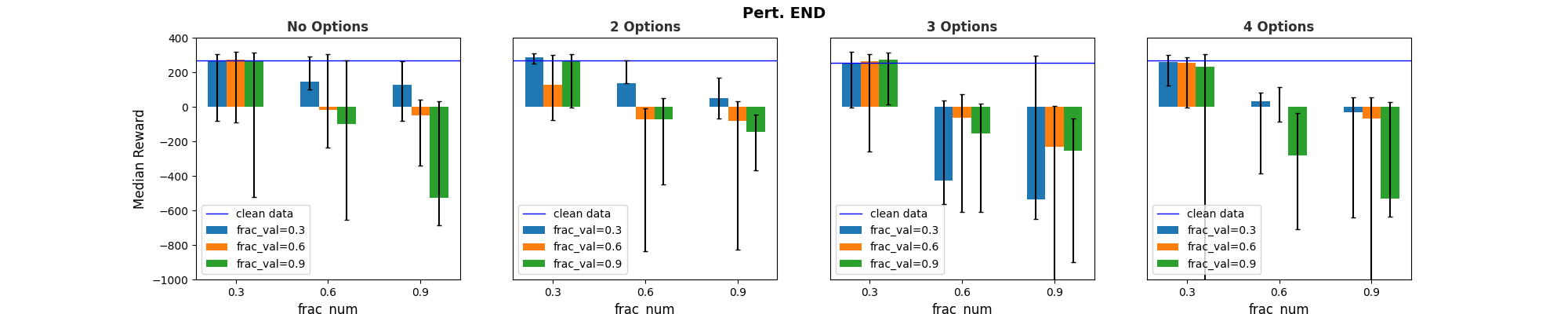}\\
\hspace*{-1.2in}
\includegraphics[width=7.5in]{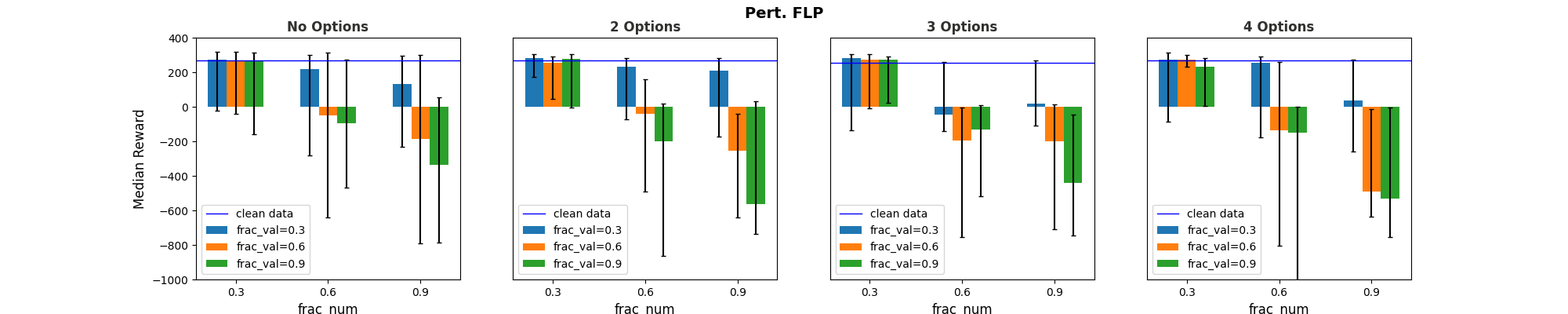}
\end{tabular}
\end{center}
\caption{Median rewards for different adversarial perturbation locations in the trajectory and different perturbation strengths.}
\label{fig_reward_perturbation}
\end{figure}
}

\subsection{Experimental Validation}
We evaluated the performance of our proposed technique using the following hypotheses:\\
\noindent
{\bf H1. Adversarial Perturbation Effect:
} Increasing the amount of perturbation in the expert demonstration trajectories decreases the performance of conventional imitation learning.\\
{\bf H2. Trajectory Accept/Reject Decision based on Trajectory Divergence:} A supervised learning based classifier that combines the occupancy measure and  Fr\'echet distance metrics of demonstrated trajectories can identify parts of the trajectories that have been adversarially modified with acceptable accuracy.\\
{\bf H3. Trajectory Repair:}  The proposed options-based, trajectory repair technique (Algorithms~\ref{algo:traj_repair}) can avoid learning from parts of demonstrated trajectories that have been adversarially modified so that the learning agent's performance does not degrade\\
\omitit{
{\bf H4. Explainability:} Using our options-based, trajectory repair technique (Algorithms~\ref{algo:traj_repair}) it is possible to determine which portions of a demonstrated trajectory cause the learning agent's performance to degrade.\\
{\bf H5. Time Overhead:} The option chaining technique (Algorithm~\ref{algo:chain-options} increases the time overhead of calculating the policy over a conventional RL-based technique by a small, acceptable amount.
}

\begin{figure}[thb!]
    \centering
    \includegraphics[width=3.5in]{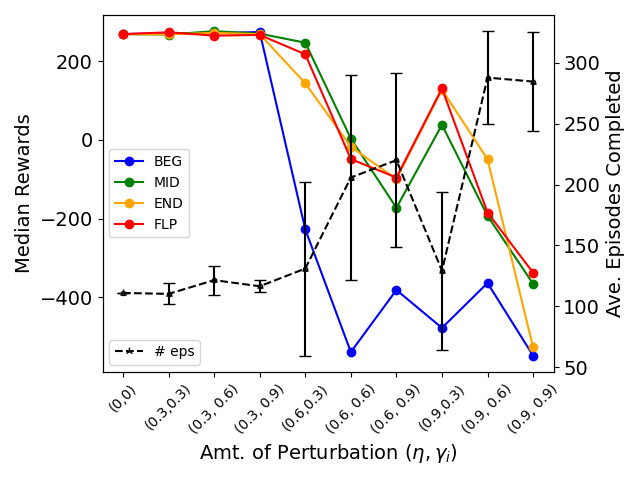}
    \caption{Change in cumulative rewards and number of episodes completed in $10,000$ time-steps of evaluation with different levels of perturbation of the expert demonstrations.}
    \label{fig:perturb_rewards_episodes_RL}
\end{figure}

To validate our Hypothesis $1$ that the strength of the adversarial perturbation in the demonstrations reduces the rewards and learning time of the learned model, we evaluated the effect of gradually increasing the number of trajectories modified ($\eta$) and the fraction of modified actions within each trajectory $\gamma_i$. While it is intuitive that increasing either $\eta$ and $\gamma_i$ will reduce the learned model's rewards. We want to understand the degree to which each of these parameters affect its performance, while applying the perturbations at different locations in demonstrated trajectories. Figure~\ref{fig:perturb_rewards_episodes_RL} shows the effect of changing the amount of perturbation in the expert demonstrations on the cumulative median rewards for different attack locations, BEG, MID and END, of the directed attack and for the gradient-based attack (FLP). We observer than when a small fraction of the expert demonstration set is changed ($\eta =0.3$), the rewards are affected nominally for all types of attacks. However, for higher values of $\eta$, the median rewards drop significantly between $-200\%$ and $-400\%$. Within a fixed value of $\eta$, $0.6$ or $0.9$, we see that changing  $\gamma_i$ (no. of actions modified inside each perturbed trajectory) also has the effect of reducing the rewards as the expert demonstrations contain more incorrect actions to learn from. We also observe that the decrease in rewards is less for attack locations MID and END, as compared to BEG for the directed attack. This makes sense because misguiding the learned model to make mistakes via demonstrating incorrect actions early on makes the trajectories veer further off from the correct course and makes it difficult for the learned model to recuperate and return on-track. Finally, we show the number and standard deviation of episodes completed (dashed line) for the different perturbation amounts, averaged over the different attack types. The number of episodes increases with increase in perturbation as with more perturbation the agent fails quickly, right after starting the task and restarts another episode, that is, there are many shorter, failed episodes with $\eta = 0.6, 0.9$ than with no or low perturbation ($\eta = 0, 0.3$). Overall, these results validate Hypothesis $1$ while showing that $\eta$ (fraction of trajectory set that is modified) has a greater effect than $\gamma_i$ (fraction of actions modified inside each modified trajectory) on the successful task completion, and, consequently, the rewards of the learned model.

\begin{figure}[tbh!]
    \centering
    \includegraphics[width=3.5in]{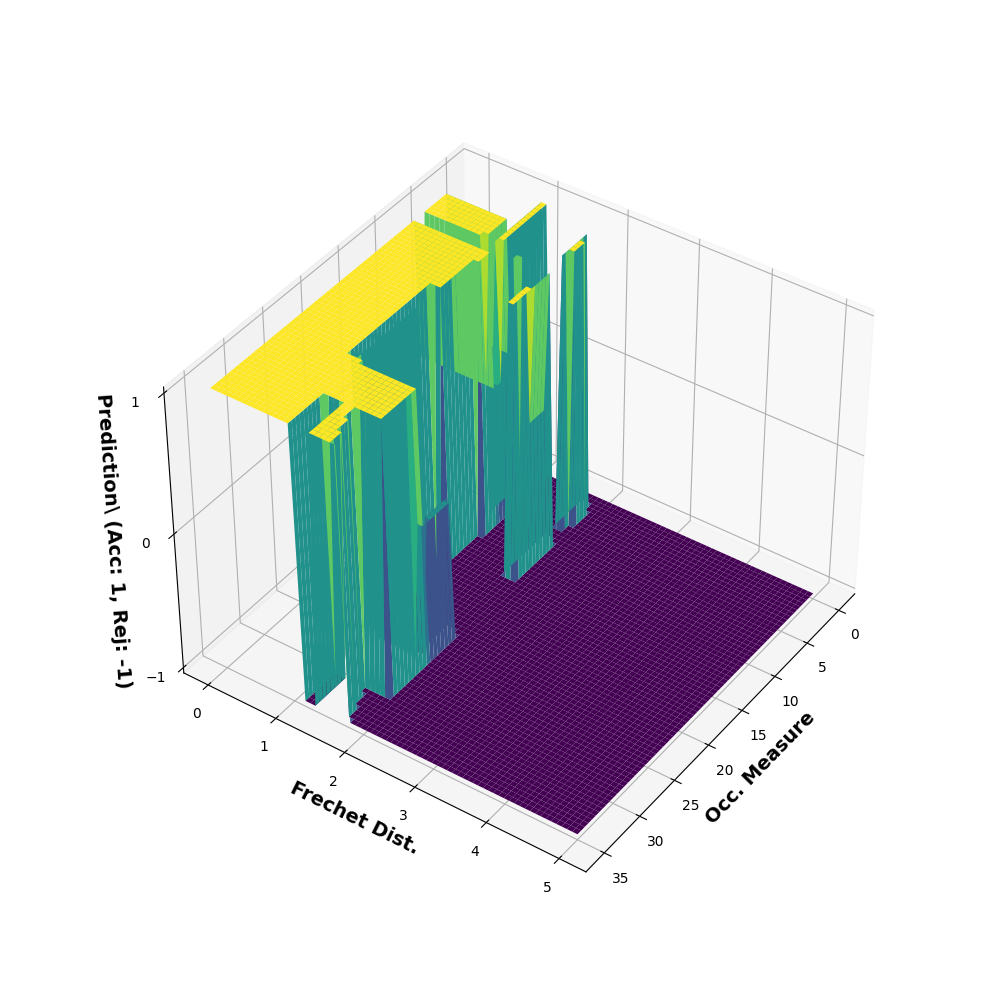}
    \caption{Boundary of the classifier for accepting (yellow surface) or rejecting (brown surface) demonstrated trajectories based on occupancy measure and Frechet distance values.}
    \label{fig:ocfd_classifer}
\end{figure}

\begin{table}[thb!]
\centering
\begin{tabular}{|c|c|c|} \hline
      & Acc. & F1\\
      \hline
     \hline
    Full traj. & $0.75$ & $0.7804$\\
    \hline
    Part traj.  &  $0.69$ & $0.8172$\\
    ($50\%$ length) & &\\
    \hline
    Part traj. & $0.71$ & $0.8098$\\
    ($33\%$ length) & &\\
    \hline
\end{tabular}
\caption{Classifier accuracy and F1 score for full and part trajectories.}%
\label{tab:ocfd_acc_f1}
\end{table}

For validating Hypothesis $2$, we trained a classifier via supervised learning and evaluated its prediction accuracy and F1-score for trajectory accept/reject decisions. For the training set of the classifier we sampled $400$ trajectories , corresponding to nearly $100,000$ state-action pairs. The training trajectories were either clean or perturbed with perturbation strengths drawn uniformly from $\eta, \gamma_i \in \{0.3, 0.6, 0.9\}$ and perturbation locations drawn uniformily from $\{$(BEG, MID, END, FLP)$\}$. Our training set is not very large{\footnote{We used $400$ trajectories in the training set as the sample diversity did not increase beyond this value for our tested LunarLander environment.}} and to improve classification accuracy with such smaller training sets, ensemble learning~\cite{zhou2012ensemble}, that combines the predictions from multiple classifiers, has been proposed as a suitable technique. We used an ensemble of classifiers with individual classifiers as: K-nearest neighbors with no. of neighbors as $2$, support vector machine with polynomial kernel function, decision tree with max-depth of $9$, and ada boost classifier with number of estimators $=50$. For the final prediction, we used ensemble voting with uniform weights given to individual classifier predictions followed by a majority voting between them. The classifier algorithms were implemented using the {\tt scikit-learn 1.2} library and the hyper-parameters in the different classifier algorithms were set to their default values given in the library. Figure~\ref{fig:ocfd_classifer} shows a profile of the learned model of classifier for different occupancy measure and Fr\'echet distance values. It indicates that the general rule learned by the classifier is to reject trajectories with very low (near zero) occupancy measure value or very high ($>\sim 1.5$) Fr\'echet distance values while for intermediate values the classification boundary exhibits a polynomial dependency on occupancy measure and Fr\'echet distance values. We tested the classifier with a test set $1000$ different trajectories that were either full length (end-to-end), or part trajectories that were either half or a third of the full length, sampled from various portions of trajectories. The classification accuracies and F1 scores for different trajectories are given in Table~\ref{tab:ocfd_acc_f1}. For all of the tested trajectories, the false negatives (accepting an adversarial trajectory) were below $10\%$. Overall, these results validate Hypothesis $2$ by showing that the classifier can be used a reliable method to identify and make accept/reject decisions for demonstrated trajectories.

\begin{table}[htb!]
    \centering
    \begin{tabular}{|c|c|c||c|c|c|c|c|}
    
    \hline
    Pert & $\eta$ & $\gamma_i$ & Full & \multicolumn{2}{c|}{Dec. w/ part} & Pre-& Post\\
    Loc. &        &            & Traj    & \multicolumn{2}{c|}{traj. (options)} & repair& repair\\
    \cline{5-6}
        &          &            &   Dec.    & Pt $1$ & Pt $2$ & reward & reward \\
    \hline    
        &      &$0.3$ & Acc & Acc$^*$ & Acc & $268.51$ & $268.51$\\
        & $0.3$&$0.6$ & Acc & Acc$^*$ & Acc$^*$ & $272.59$ & $272.59$\\
        &      &$0.9$ & Acc & Acc$^*$ & Acc$^*$ & $269.06$ & $269.06$\\
        \cline{2-8}
        &      &$0.3$ & Rej & Acc & Rej & $144.33$ & $277.37$\\
    END & $0.6$&$0.6$ & Rej & Acc & Rej & $-17.15$ & $269.69$\\
        &      &$0.9$ & Rej & Rej & Rej & $-99.89$ & $268.76$\\
        \cline{2-8}
        &      &$0.3$ & Rej & Acc & Rej & $128.38$ & $276.33$\\
        & $0.9$&$0.6$ & Rej & Acc & Rej & $-49.13$ & $261.75$\\
        &      &$0.9$ & Rej & Rej & Rej & $-528.20$& $268.76$\\
    \hline
        &      &$0.3$ & Acc & Acc$^*$ & Acc & $271.92$ & $271.92$\\
        & $0.3$&$0.6$ & Acc & Acc$^*$ & Acc & $271.79$ & $271.79$\\
        &      &$0.9$ & Acc & Acc$^*$ & Acc & $274.68$ & $274.68$\\
        \cline{2-8}
        &      &$0.3$ & Rej & Rej & Rej & $-227.53$ & $268.76$\\
    BEG & $0.6$&$0.6$ & Rej & Rej & Rej & $-539.82$ & $268.76$\\
        &      &$0.9$ & Rej & Rej & Rej & $-381.87$ & $268.76$\\
        \cline{2-8} 
        &      &$0.3$ & Rej & Rej & Rej & $-478.36$ & $268.76$\\
        & $0.9$&$0.6$ & Rej & Rej & Rej & $-364.10$ & $268.76$\\
        &      &$0.9$ & Rej & Rej & Rej & $-548.87$ & $268.76$\\
    \hline
    \hline
        &      &$0.3$ & Acc & Acc$^*$ & Acc & $273.99$ & $273.99$\\
        & $0.3$&$0.6$ & Acc & Acc$^*$ & Acc & $265.50$ & $265.50$\\
        &      &$0.9$ & Acc & Acc$^*$ & Acc & $267.33$ & $267.33$\\
        \cline{2-8}
        &      &$0.3$ & Acc & Rej & Acc & $218.05$ & $276.5$\\
    FLP & $0.6$&$0.6$ & Rej & Rej & Rej & $-48.74$ & $268.76$\\
        &      &$0.9$ & Rej & Rej & Rej & $-95.63$ & $268.76$\\
        \cline{2-8}
        &      &$0.3$ & Acc & Rej & Acc & $131.03$ & $261.02$\\
        & $0.9$&$0.6$ & Rej & Rej & Rej & $-184.74$ & $268.76$\\
        &      &$0.9$ & Rej & Rej & Rej & $-337.80$ & $268.76$\\
    \hline
    \end{tabular}
    \caption{Accept/reject decisions and median rewards before and after repair for trajectories divided into $2$ parts and modified by directed attack at locations BEG and END, and gradient-based attack $FLP$. Asterisks mark decisions where classifier predicted 'Reject', but return ratio condition (Line 7, Algo.~\ref{algo:traj_repair} changed to 'Accept'.}
    \label{tab:repair_2_parts}
\end{table}

For validating Hypothesis 3, on trajectory repair, we experimented with $2$ and $3$ partitions of the trajectory. For each partition, we considered that the adversary had perturbed the trajectory at different locations BEG, END and FLP. Table~\ref{tab:repair_2_parts} shows the results for trajectory repair using our proposed technique when trajectories are divided into two equal parts, including the decision made by the trajectory accept/reject classifier for full and two part trajectories (with and without options), and the rewards before and after trajectory repair. Note that the first three columns of Table~\ref{tab:repair_2_parts}, perturb location, $\eta$ and $\gamma_i$ are given for legibility and not known by the learning agent or repair technique. The results show that for all cases adversarially modified trajectories can be identified and repaired at the portions that were modified, while preserving the portions that were not modified, and preventing the performance of the learning agent from degrading as shown by the median rewards similar to the clean trajectory reward values. The main impact of our trajectory repair technique is seen for perturb location END, as part trajectories are repaired to improve the reward to values similar to clean trajectory rewards. Moreover, using trajectory repair, we are able to detect and use the clean part of the trajectory, thereby improving sample efficiency. For perturbation location BEG, trajectories are mostly rejected because, as the decisions are sequential, modifying actions early on in an episode result in incorrect or sub-optimal actions downstream in the episode. For $\eta=0.3$, for all perturbation locations, we see that some part trajectories get rejected by the classifier when not using the return ratio condition in Line 7 of Algorithm~\ref{algo:traj_repair} (marked with asterisk in Table~\ref{tab:repair_2_parts}. This happens because for our {\tt LunarLander} task, different episodes start from different initial locations and their occupancy measure and Fr\'etchet distance values show a larger divergence in the initial part of episodes. However, for all these cases, we note that the reward is not degraded using the return ratio condition. The gradient-based attack, FLP, is more difficult to detect as the perturbation locations made the adversary in the trajectories are selected strategically and are not successive, as in the directed attack. The bottom part of Table~\ref{tab:repair_2_parts} shows that our technique works successfully for gradient-based attacks as well and is able to discern and reject modified trajectories and learn only from the clean parts of trajectories, when available. Our experiments with perturb location MID (not reported here) showed similar results as BEG and END - parts of trajectories before the perturb location were accepted by the classifier while those following the perturb location were rejected as they were downstream and affected by the perturbation; rewards in all cases were restored to values similar to clean trajectories following trajectory repair.

Table~\ref{tab:repair_3_parts} shows the trajectory repair results when the trajectories are divided into three parts. We report the results for gradient-based (FLP) attacks only as they are more difficult to detect. Here too, we see that the trajectory repair technique is able to identify parts of trajectories that have lower perturbation and can be used for learning without degrading the task performance, as shown by the restored reward values for these cases.

\begin{table}[]
    \centering
    \begin{tabular}{|c|c|c||c|c|c|c|c|c|}
    
    \hline
    Pert & $\eta$ & $\gamma_i$ & Full & \multicolumn{3}{c|}{Dec. w/ part} & Reward &  Reward \\
    Loc. &        &            & Traj.    & \multicolumn{3}{c|}{traj. (options)} & before & post-\\
    \cline{5-7}
        &          &            &  Dec.    & Pt $1$ & Pt $2$ & Pt $3$ & repair & repair\\
    \hline    
    \hline
        &      &$0.3$ & Acc & Acc$^*$ & Acc$^*$ & Acc & $279.63$ & $279.63$ \\
        & $0.3$&$0.6$ & Acc & Acc$^*$ & Acc$^*$ & Acc & $274.76$ & $274.76$ \\
        &      &$0.9$ & Acc & Acc$^*$ & Acc & Acc & $274.82$ & $274.82$\\
        \cline{2-9}
        &      &$0.3$ & Acc & Acc & Rej & Rej & $-45.05$ & $271.09$\\
    FLP & $0.6$&$0.6$ & Rej & Rej & Rej & Rej & $-193.76$ & $255.52$\\
        &      &$0.9$ & Rej & Rej & Rej & Rej & $-133.92$ & $255.52$\\
        \cline{2-9}
        &      &$0.3$ & Rej & Acc & Rej & Rej & $19.30$ & $275.60$ \\
        & $0.9$&$0.6$ & Rej & Rej & Rej & Rej & $-198.05$ & $255.52$\\
        &      &$0.9$ & Rej & Rej & Rej & Rej & $-439.68$ & $255.52$\\
    \hline
    \end{tabular}
    \caption{Accept/reject decisions and median rewards before and after repair for 3-part trajectories modified by the gradient-based attack (FLP). As before, asterisks mark decisions where classifier 'Reject' decision is overridden by return ratio condition.}
    \label{tab:repair_3_parts}
\end{table}

\subsection{Ablation Experiments}

\begin{figure}[thb!]
    \centering
    \includegraphics[width=3.0in]{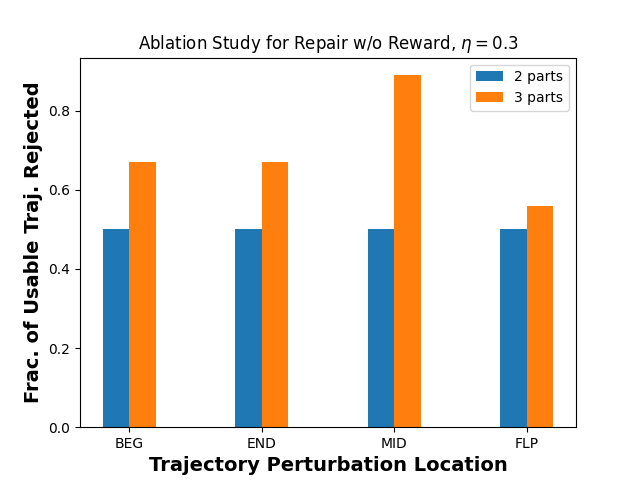}
    \caption{Fraction of usable trajectories rejected by classifier without using the reward ratio between demonstrated and clean trajectories in Algorithm~\ref{algo:traj_repair} when $\eta=0.3$. For all other values of $\eta$, no usable trajectories get rejected by the classifier without using the reward ratio.}
    \label{fig:ablations_decisions_without_reward_repair}
\end{figure}

We performed two ablation experiments by removing certain features of our algorithms to understand their effect on the results
. 
{\bf Effect of Return Ratio Condition.} In the first experiment, we tested the effect of using the return ratio condition to override a reject decision made by the trajectory classifier ( Line $7$, Algorithm~\ref{algo:traj_repair}). The reward ratio is provided as a guard rail against false positives from the classifier so that correct, benign trajectories that show a new way to peform the task and have a higher divergence measure from known, clean trajectories do not get discarded. For this experiment, we varied the perturbation strengths, $\eta=\{0.3, 0.6, 0.9\}$, the perturbation locations, $\{\text{BEG, MID, END, FLP}\}$, and the number of trajectory parts, $\{2, 3\}$, and recorded the average fraction of trajectories that got changed from 'Accept' to 'Reject' when not using the reward ratio condition. For $\eta=\{0.6, 0.9\}$, for all perturbation location, we observed that none of the classifier decisions were changed after removing the reward ratio, for both $2-$ and $3-$ part trajectories. This indicates that for higher perturbation strengths, false positives are absent or rare and the reward ratio condition is not triggered. For $\eta=0.3$, our results are shown in Figure~\ref{fig:ablations_decisions_without_reward_repair}. We see that for $2-$ part trajectories $50\%$ of the trajectory is discarded, while for $3-$ part trajectories between $50-90\%$ of the trajectory gets discarded. However, although discarding trajectories deteriorates sample efficiency, it does not affect the learning performance as the difference in the rewards with and without the reward ratio was nominal (within $1-3\%$). In general, our findings from this experiment indicate that when the perturbation strength is low and the divergence measure has difficulty in classifying a trajectory accept/reject decision, the reward ratio condition is important to prevent valid but divergent trajectories from getting discarded. 

\begin{figure}[thb!]
    \centering
    \begin{tabular}{cc}
        \includegraphics[width=2.7in]{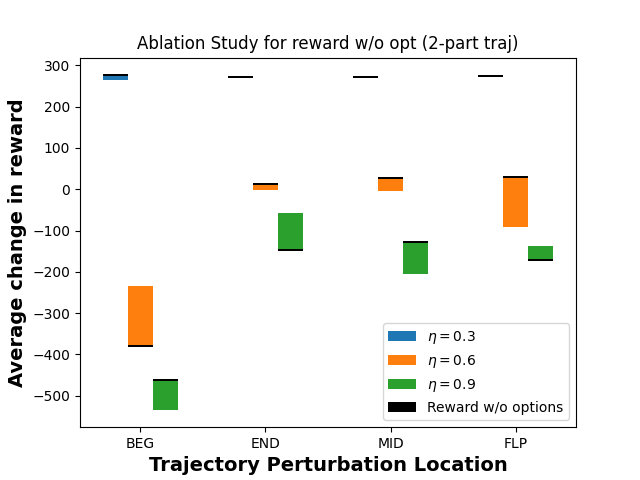} &  
        \includegraphics[width=2.7in]{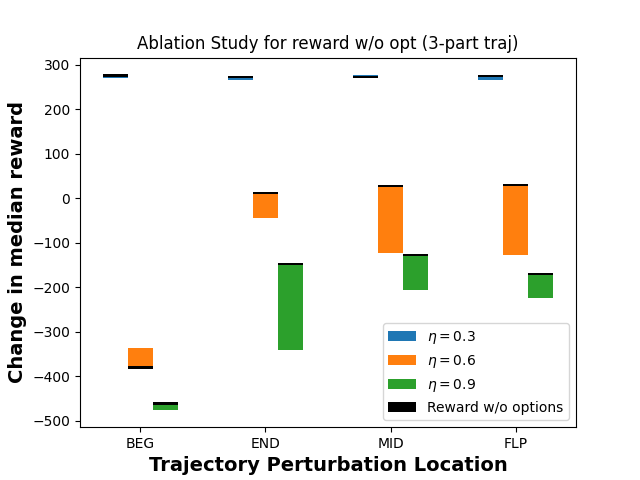}  \\
        {\small (a)} & {\small (b)}
    \end{tabular}
    \caption{Ablation experiments for overhead of using options showing the change in median reward with and without options for different perturbation strengths ($\eta=\{0.3, 0.6, 0.9\})$ and different perturbation locations, $\{\text{BEG, END, MID, FLP}\}$, black lines at one end of the bars denote the reward without options. (a) $2$-part trajectories, (b) $3$-part trajectories.}
    \label{fig:ablations_options}
\end{figure}

{\bf Overhead introduced by Options.} Options are the key component of our technique as they facilitate partitioning trajectories and retaining and learning from only the usable part trajectories. To determine the feasibility of our technique there are two important questions related to using options that need to be addressed: does using options affect the learning performance in terms of rewards and can options be used without degrading the rewards?  To answer these questions, we performed our next ablation experiment. We trained the agent to learn to play the {\tt LunarLander} game using full trajectories versus using part trajectories via options, and recorded the difference in median rewards for these two settings for different perturbation strengths, $\eta=\{0.3, 0.6, 0.9\}$, different perturbation locations, $\{\text{BEG, MID, END, FLP}\}$, and different number of trajectory parts, $\{2, 3\}$. Our results are shown in Figure~\ref{fig:ablations_options}, the black lines at one end of the bars show the reward without options and the bars show the change in rewards using options. We see that when there is little perturbation ($\eta=0,3$), using options has negligible change in rewards, around $<1-2\%$. When the perturbation increases to $\eta=\{0.6, 0.9\}$, the rewards for using options either increases or decreases from the reward without options. This indicates that perturbed trajectories make it difficult to chain options. Also, as chaining options has to be done between every pair of trajectory parts, as the number of trajectory parts increases, the decrease in rewards from using options also becomes more pronounced. However, when options are repaired using Algorithm~\ref{algo:traj_repair}, the part trajectories can again be chained efficiently and the rewards are again restored to higher values, similar to those learned for clean trajectories. Overall, these experiments show that our approach of repairing part trajectories with options does not introduce significant overhead in the computations of the imitation learning algorithm.

\section{Conclusions and Future Work}
In this paper, we proposed a novel technique using options to selectively include portions of demonstrated trajectories for training the policy of an imitation learning-based agent in the presence of demonstrations given by potentially adversarial experts. Our results show that using our technique, the learned policy can prevent learning from portions of trajectories that would degrade the agent's reward. Our technique provides two main advantages: it improves the robustness of the policy training as well as the sample complexity of the demonstration samples without resulting in a significant overhead of the policy training time. Closely related to our research is the field of opponent modeling, cross-play and inter-play where agents build models of their opponents' behaviors from observations and train their policies by playing against those models. A potential problem in opponent modeling is deception by opponents where opponents can demonstrate incorrect behaviors via trajectories to misguide an agent. Our proposed trajectory repair technique could be used in such situations to identify deceptive trajectories by comparing them with trajectories of known or rational opponent behaviors and prevent learned policies from getting misled.

One of the requirements in our technique is that it requires a human to identify a base set of clean policies with which the agent's task is performed successfully. While most real-life domains require human subject matter experts to provide such feedback, techniques like inverse reinforcement learning that automatically update the reward function to improve the agent's performance could be used to reduce the technique's reliance on human expertise. Another aspect of our work is that it assumes that clean trajectories have low divergence between them, For tasks that can be solved in different ways, clean trajectories representing different ways to solve the task might have high divergence with each other. In such cases, the different ways of performing the task could be grouped into clean trajectory clusters, and the divergence with demonstrated trajectories could be determined for clean trajectory clusters to make the accept/reject decision for our technique.

We have used behavior cloning as our imitation learning algorithm. More sophisticated imitation learning algorithms like the data aggregation ({\tt DAgger}) algorithm~\cite{ross2011reduction} or the generative adversarial imitation learning from options (GAIfO)~\cite{TorabiWS19a} could be used in place of behavior cloning. These algorithms are likely to improve the performance of the imitation learning portion, and our proposed options-based technique could still be used with them to partition demonstrated trajectories and identify acceptable partial trajectories for learning. 

Our proposed technique was aimed at enabling an agent to use imitation learning in the presence of adversarial trajectories. It is likely that smart adversaries will discover that its attacks are not effective against a learning agent that has used our technique to avoid accepting adversarial trajectories. It could then craft new types of adversarial trajectory attacks to evade the trajectory accept/reject decision classifier. Such situations could be modeled as a higher-level, adversarial game between the adversary and the learning agent and techniques from hierarchical reinforcement learning~\cite{hutsebaut2022hierarchical} and Bayesian games~\cite{meta2022human} could be used to solve them.

We envisage that further investigation of the options based technique for adversarial imitation learning described in this paper will lead to new insights into the problem of learning for demonstrations and could be used by a learning agent to quickly and robustly learn effective operations in new, open environments from clean as well as adversarial trajectories. 

\section*{Acknowledgements}
This work was supported by the U.S. Office of Naval Research as part of the FY$21$ NRL Base Funding $6.1$ project Game Theoretic Machine Learning for Defense Applications.

\bibliographystyle{abbrv}
\bibliography{refs}

\end{document}